\documentclass[sigconf]{acmart}
\usepackage{epsfig}
\usepackage[ruled]{algorithm2e}
\usepackage{threeparttable}
\usepackage{adjustbox}
\usepackage{url}
\usepackage{booktabs}

\usepackage{amsmath,amsfonts,amscd,amsthm,xspace}
\usepackage{stfloats}
\usepackage{balance}
\usepackage{tabularx}
\usepackage{multirow, multicol}

\usepackage{caption}
\usepackage{subcaption}

\SetKwProg{Fn}{function}{}{}

\copyrightyear{2021}
\acmYear{2021}
\setcopyright{acmlicensed}\acmConference[SAC '21]{The 36th ACM/SIGAPP Symposium on Applied Computing}{March 22--26, 2021}{Virtual Event, Republic of Korea}
\acmBooktitle{The 36th ACM/SIGAPP Symposium on Applied Computing (SAC '21), March 22--26, 2021, Virtual Event, Republic of Korea}
\acmPrice{15.00}
\acmDOI{10.1145/3412841.3441958}
\acmISBN{978-1-4503-8104-8/21/03}

\begin{document}

\title[Data Augmentation in a Hybrid Approach for ABSA]{Data Augmentation in a Hybrid Approach for Aspect-Based Sentiment Analysis}

\author{Tomas Liesting}
\affiliation{
  \institution{Erasmus University Rotterdam}
  \streetaddress{Burgemeester Oudlaan 50}
  \city{Rotterdam}
  \country{the Netherlands}
  \postcode{3062 PA}
}
\email{tomas.liesting@gmail.com}

\author{Flavius Frasincar}
\orcid{0000-0002-8031-758X}
\affiliation{  \institution{Erasmus University Rotterdam}
  \streetaddress{Burgemeester Oudlaan 50}
  \city{Rotterdam}
  \country{the Netherlands}
  \postcode{3062 PA}
}
\email{frasincar@ese.eur.nl}

\author{Maria Mihaela Tru\c{s}c\u{a}}
\affiliation{  \institution{Bucharest Univ. of Economic Studies}
  \city{Bucharest}
  \country{Romania}
  \postcode{010374}
  \streetaddress{Piata Romana 6}
}
\email{maria.trusca@csie.ase.ro}

\begin{abstract}
Data augmentation is a way to increase the diversity of available data by applying constrained transformations on the original data. This strategy has been widely used in image classification but has to the best of our knowledge not yet been used in aspect-based sentiment analysis (ABSA). ABSA is a text analysis technique that determines aspects and their associated sentiment in opinionated text. In this paper, we investigate the effect of data augmentation on a state-of-the-art hybrid approach for aspect-based sentiment analysis (HAABSA). We apply modified versions of easy data augmentation (EDA), backtranslation, and word mixup. We evaluate the proposed techniques on the SemEval 2015 and SemEval 2016 datasets. The best result is obtained with the adjusted version of EDA, which yields a $0.5$ percentage point improvement on the SemEval 2016 dataset and $1$ percentage point increase on the SemEval 2015 dataset compared to the original HAABSA model. 
\end{abstract}

%



\maketitle


\section{Introduction}
Nowadays, there are many online platforms to compare and review products, locations, and services. Research has demonstrated that 91\% of the people regularly read these online reviews, and around 84\% trust them \citep{craig2017percent}, implying that these opinionated texts influence the decisions of the majority of the consumers. Therefore, it is not surprising that the analysis of these opinionated texts awoke much interest in the past years. As the available information is enormous (e.g., Tripadvisor by itself already has over 435 million reviews \citep{tripadvisor2017network}), the manual analysis is nearly impossible, meaning that the interest in automatic sentiment analysis has increased a lot as well. 


In this research, the focus lies on aspect-based sentiment analysis (ABSA), where the goal is to predict the sentiment of a certain aspect \citep{schouten2016survey}. ABSA has three crucial components. First, the target words that the sentiment is about have to be extracted (target extraction). Second, the aspects have to be detected depending on the discussed product or service (aspect detection). The third step is sentiment classification, in which the polarity of the opinion is determined (e.g., positive, negative, or neutral).


Currently, almost all state-of-the-art (SOTA) methods in ABSA rely on supervised learning due to its high rates of effectiveness. However, the major downside of supervised training is the unavailability of large annotated datasets. To mitigate this problem many researchers opt for transfer learning from other similar knowledge domains. As this approach might be too costly because it requires the access to external resources, another solution to extend the current training dataset is by using data augmentation, mainly employed in analyzing images \citep{perez2017effectiveness}. The idea of data augmentation is increasing the available information by doing semantically constrained transformations on the training data. The idea is intuitive in image classification, as using the same image but shifted, zoomed, rotated, cropped, and many other transformations lead to more information improving performance \citep{perez2017effectiveness}, but can also be applied in natural language processing (NLP) \cite{wei2019eda}.


In this paper, we explore how data augmentation techniques can improve the sentiment classification task. For this, we are not only interested in extending the training set but also in keeping a good quality of it. The baseline model we choose to evaluate our data augmentation techniques achieves high accuracy in sentiment classification and is proposed in \citep{HAABSA} as a hybrid model with two steps. First, an ontology-based reasoner attempts to determine the polarity of a given aspect in text. If the ontology is not conclusive, the task is solved by a neural network (NN). The backup model is based on the LCR-Rot neural network \citep{Zheng2018lcrrot}. The model splits the sentences into three parts, left context, target, and right context, used as input for three LSTMS cells and a rotatory attention. The model furthermore uses word embeddings from GloVe \citep{pennington2014glove} and obtains SOTA results for several ABSA sentiment evaluation tasks. The source code of our work can be found on GitHub at \url{https://github.com/tomasLiesting/HAABSADA}.

The contribution of our paper can be summarised as follows:
\begin{enumerate}
    \item We introduce an NLP data augmentation framework together with a set of variations suitable for the sentiment classification task.
    \item Taking as example two widely used ABSA datasets, we prove that our techniques can boost the accuracy of the baseline model up to 2 percentage points.
\end{enumerate}



The paper is organized as follows. In Section \ref{sec:literature}, relevant data augmentation techniques for NLP used by other researchers are presented. In Section \ref{sec:data} the datasets used for training and evaluation are introduced. Afterward, in Section \ref{sec:methodology}, we explain how the techniques discussed in Section \ref{sec:literature} can be used for ABSA, and we propose several extensions of these. In Section \ref{sec:evaluation}, the effect of these techniques on the accuracy of the model is given and discussed. In Section \ref{sec:concludingremarks}, we present the implications of our research and propose ideas for future research.

\section{Related Work} \label{sec:literature}

Data augmentation increases the number of data points in a training set by transforming data in a constrained manner. This technique has been prevalent in image classification, where its effectiveness has already been proven (e.g., by \citet{perez2017effectiveness}). In \cite{perez2017effectiveness}, the authors evaluate several techniques where a given image is rotated, tilted, cropped, or shifted, improving predictions. In NLP, some attention has been given to data augmentation as well, which is further discussed in this section.

First of all, \citet{wei2019eda} create a generalized way to augment textual data. They propose easy data augmentation (EDA), which consists of four different data augmentation methods in NLP. First, they use Synonym Replacement (SR), where $n$ words, which are not stopwords, are selected and replaced with synonyms. Secondly, they use Random Insertion, where they insert a random synonym of a word at a random position in a sentence. Their third method is Random Swap, where two words in a sentence are randomly swapped, and lastly, they propose Random Deletion, where random words within the sentence are deleted. They find that using 50\% of the original data, they generally obtain similar results to the models not using data augmentation and that they achieve better results in all the tasks on which the data augmentation task has been performed using the full dataset. They do note that the data augmentation effect is more substantial on smaller datasets in comparison to larger datasets.

Another way to use data augmentation is introduced by \citet{sennrich2016improving}, employed initially to improve machine translation. They translate the data into another language and afterward translate it back to obtain synthetic data. This technique is called backtranslation. They show that this method obtains better results in translation tasks. Though no research has been done on the use of backtranslation in ABSA, the method shows potential for NLP in general. \citet{yu2018fast} used backtranslation from English to French or German and back in order to enhance their dataset. This backtranslation initially resulted in twice as much data (every sentence is translated to French and back to English and is treated as a new sentence), which yielded an improvement on F1-scores of 0.5 percentage points on the SQuAD dataset \citep{squad}. When also using German backtranslation, obtaining three times as much data, this method resulted in another increase of 0.2 percentage points on top of the 0.5 percentage points. The authors also notice that augmenting data more than three times results in decreased performance, presumably because this backtranslated data is noisy compared to the original data. They modified the ratio between original and augmented data, and empirically found that a 3:1:1 ratio (three times the original data, one time the data backtranslated from French and one time the data backtranslated from German) resulted in the highest performance gain of 1.1-1.5 percent. This means that they used five times as much data for their best result.

\citet{schleifer2019low} used a combination of the previous methods on a pretrained multilayered AWD-LSTM model \citep{merity2017regularizing} on an IMDB movie dataset. In this analysis, the authors found that using EDA on the full dataset does not lead to substantial improvements on test classification, and that backtranslation has a small gain. Using a subset of the dataset does lead to improved performances with backtranslation. \citet{wei2019eda} suspect that other models that use word embeddings like BERT will not benefit either from the use of EDA, as the word embeddings are contextual. 

The last method used for data augmentation is the mixup, originally introduced by \citet{zhang2017mixup} for image recognition. Their idea is to (linearly) interpolate between feature vectors of an image, which should be identical to the interpolation of the associated target vectors. The authors take two images and their corresponding target, $(x_i; y_i)$ and $(x_j; y_j)$, where $x$ and $y$ are the image and the target, respectively, and a value $\lambda$ is drawn from the distribution $Beta(\alpha, \beta)$ distribution, where the authors set $\alpha = \beta$ as this yields a symmetric distribution, and $\alpha \in [0.1, 0.4]$, as higher values resulted in underfitting. They create a synthetic image $(\tilde{x}_{ij}; \tilde{y}_{ij})$ as given below:
\begin{equation}
\label{mixup_images}
\begin{split}
    \tilde{x}_{ij} = \lambda x_i + (1-\lambda) x_j \\
    \tilde{y}_{ij} = \lambda y_i + (1-\lambda) y_j
\end{split}
\end{equation}

As word embeddings are essentially feature vectors, similar to images, the mixup can also be applied to NLP tasks \citep{ghou2019mixup}. In  \cite{ghou2019mixup}, the proposed adaptation for NLP is twofold. First, individual word embeddings can be interpolated. This is done by zero-padding the sentences to make them of the same length, after which interpolation is done on every word in the sentence. This means that the first word of sentence A is interpolated with the first word of sentence B, and so forth. Second, the model uses sentence embeddings. Here the sentences are fed to the model and are encoded by an LSTM or CNN into sentence embeddings. These representations are extracted afterward and a linear interpolation is performed on two representations. These two methods have shown to improve the accuracy of sentence classification with both CNN and LSTM significantly.

To summarize, the usage of EDA as introduced by \citet{wei2019eda} is an interesting development, though the effect of this method appears to be small when using (contextual) word embeddings. Backtranslation has proven to provide small to substantial improvements on large and small datasets, respectively. Word and sentence mixups appear to be interesting developments in both large and small datasets. In the field of ABSA, data augmentation has not yet been used to the best of our knowledge.

\section{Data}\label{sec:data}
The used data is given by the SemEval (or Semantic Evaluation) datasets of 2015 and 2016. SemEval is a series of evaluation workshops that aims to extract meanings out of sentences. The datasets are annotated by human linguists, and NLP algorithms aim to be as close to the human annotator as possible. SemEval datasets have different tasks. In our case, we use task 12 subtask 1 of SemEval 2015 \citep{pontiki2015semeval} and task 5 subtask 2 of SemEval 2016 \citep{pontiki2016semeval}. The goal of these tasks is to predict the polarity of a sentence about a given aspect. This way, the performance benefits when applying the data augmentation can be measured.

\begin{figure}[h] 
	\centerline{\psfig{figure=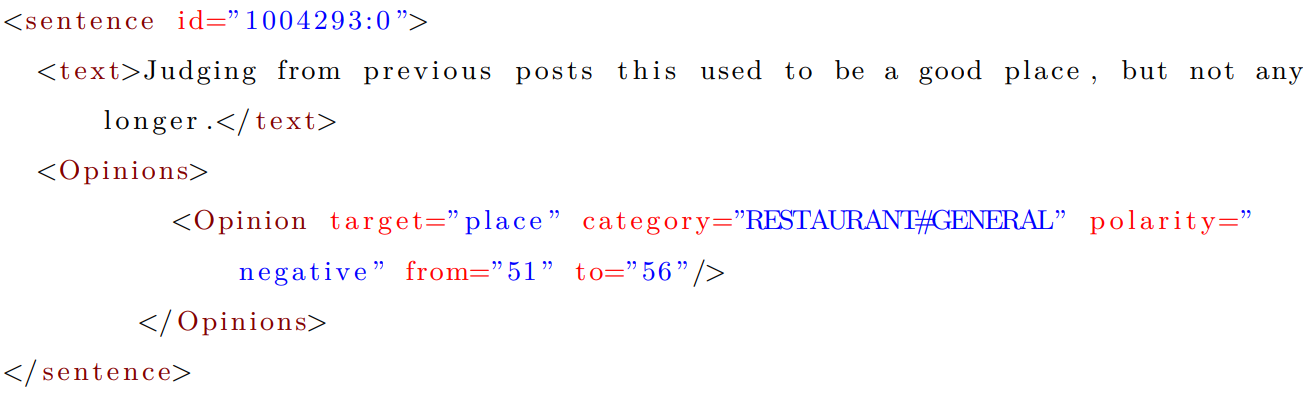,width=86.7mm}}
    \caption{Annotated data fragment}
    \label{fig:data_sample}
\end{figure}

The previously considered two datasets contain restaurant reviews with one to several sentences represented in the XML format. In each sentence, an aspect, a category, and a polarity are given. An example of a sentence in the dataset from SemEval 2016 can be seen in Figure \ref{fig:data_sample}.  In this figure the aspect is given in the \textit{target} field, a general category is given in the \textit{category} field, and the polarity of the sentiment with respect to the previous category is given in the \textit{polarity} field. The dataset of SemEval 2015 has a similar structure. The preprocessing of the data and the computation of word embeddings are done similarly to the work of \citet{HAABSA}.

Table \ref{tab:polarities} shows the frequencies of polarities for the training and testing data of the employed datasets. In all training sets, neutral sentences are the least frequent, and positive are the most frequent. Table \ref{tab:categories} gives an overview of the categories in the SemEval datasets. Food quality, general service, and general ambiance are the most represented categories, making up well over 50 percent of the data. The relative frequency of the categories in the test and train data appears similar for both datasets, though the ratio is more similar in 2016 than in 2015. This is also the case for the polarities. 

\begin{table*}[t] 
\caption{Frequencies of polarities in the SemEval 2015 and 2016 train and test datasets.}
\label{tab:polarities}
\centering
\small
\begin{tabular}{l|ll|ll|ll|ll} 
\multirow{2}{*}{Dataset} & \multicolumn{2}{c|}{Positive} & \multicolumn{2}{c|}{Neutral} & \multicolumn{2}{c|}{Negative} & \multicolumn{2}{c}{Total} \\
& Frequency & \% & Frequency & \% & Frequency & \% & Frequency & \% \\ 
\hline
SemEval 2015 train & 963 & 75 & 36 & 3 & 280 & 22 & 1279 & 100  \\
SemEval 2015 test  & 353 & 59 & 37 & 6 & 207 & 35 & 597 & 100  \\
SemEval 2016 train & 1319 & 70 & 72 & 4 & 488 & 26 & 1879 & 100  \\
SemEval 2016 test  & 483 & 74 & 32 & 5 & 135 & 21 & 650 & 100  \\ 

\end{tabular}
\end{table*}

\begin{table}[] 
\caption{Categories in the SemEval 2015 and 2016 train and test datasets.}
\label{tab:categories}
\centering
\small
\begin{tabular}{l|ll|ll} 
\multirow{2}{*}{{Category}} & \multicolumn{2}{c|}{2015} & \multicolumn{2}{c}{2016}  \\
& train & test & train & test \\ 
\hline
FOOD\#QUALITY     & 524 & 242 & 765 & 283   \\
SERVICE\#GENERAL  & 217 & 104 & 324 & 107   \\
AMBIANCE\#GENERAL & 164 & 68 & 228 &  59 \\
RESTAURANT\#GENERAL  & 124 & 59 & 183 & 58 \\ 
FOOD\#STYLE\_OPTIONS & 81 & 33 & 116 & 51\\
FOOD\#PRICES & 41 & 26 & 71 & 22 \\
DRINKS\#QUALITY & 32 & 11 & 44 & 22 \\
RESTAURANT\#MISCELLANEOUS & 30 & 19 & 49 & 18 \\
DRINKS\#STYLE\_OPTIONS & 26 & 6 & 32 & 11 \\
LOCATION\#GENERAL & 14 & 8 & 22 & 10 \\
RESTAURANT\#PRICES & 10 & 16 & 26 & 5 \\
DRINKS\#PRICES & 15 & 5 & 20 & 4 \\
FOOD\#GENERAL & 1 & 0 & 0 & 0 \\
\end{tabular}
\end{table}

\section{Framework}\label{sec:methodology}
This section gives an overview of the framework used in this paper. Section \ref{sec:ontology} gives a short presentation of the ontology reasoner used in HAABSA. Afterward, Section \ref{sec:lcrrot} gives a description of the used machine learning method and the used word embeddings. Last, Section \ref{sec:dataaugmentation} explains existing data augmentation techniques and proposed extensions appropriate for ABSA. 

\subsection{Ontology Reasoner} \label{sec:ontology}
HAABSA is a hybrid model for ABSA that consists of two stages. First of all, an ontology reasoner is employed, which is similar to the reasoner used by \citet{schouten2018ontology}. This reasoner uses a domain-specific sentiment ontology to determine the polarity of a sentence about an aspect. The domain sentiment ontology groups concepts in three classes: \textit{SentimentValue}, \textit{Aspect Mention}, and \textit{SentimentMention}. The first contains two classes, namely \textit{Positive} and \textit{Negative}. The \textit{AspectMention} provides lexical representation for aspect categories. An example of this is the word \textit{fish} which is linked to the category \textit{FOOD\#QUALITY}. 

The \textit{SentimentMention} class determines whether the sentiment of a sentiment expression is positive or negative for a specific aspect. There are three types of \textit{SentimentMentions}. Type 1 concerns words that have the same sentiment for every aspect. An example is the word \textit{bad}, which is always negative, no matter the context. Type 2 are words that always have the same polarity but only apply to some aspects. An example is the word \textit{delicious}, which can be applied to food and drinks, but not, for example, to a couch. Type 3 is the last one and represents words of which their polarity is dependent on the context. Take, for example, the word \textit{cold}. If ice-cream is cold, it has a positive or neutral polarity, while cold fries generally have a negative polarity.

\subsection{Multi-Hop Left-Center-Right Neural Network with Rotatory Attention} \label{sec:lcrrot}
The second step of the hybrid model is the machine learning method. As the ontology reasoner can only predict sentiment in 60\% of the cases, an ML technique is used to classify the remaining data \citep{schouten2018ontology}. The best model presented by \citet{HAABSA}, the multi-hop LCR-Rot, splits up a sentence into left context, target, and right context vectors. These vectors are the input for a left, center, and right bidirectional LSTM cell, respectively. The outputs of these cells are the inputs for a rotatory attention mechanism, consisting of two steps. The first step determines the most indicative words in the left and right contexts, and the second step captures the most important words for the target. The rotatory attention is applied for several iterations, where three iterations yielded the best results.

For the ML technique, the word embeddings are created using the GloVe (or global vectors) embeddings \citep{pennington2014glove}. The advantage of GloVe is that, unlike the skip-gram or CBOW methods, GloVe uses both local information and a global word co-occurrence matrix to create word embeddings. This method resulted in the best performance of the hybrid model compared to the CBOW and skip-gram methods. The word embeddings of GloVe are freely available for download\footnote{http://nlp.stanford.edu/data/glove.42B.300d.zip}.

\subsection{Data Augmentation} \label{sec:dataaugmentation}
Data augmentation in ABSA seems to be an untouched subject. For NLP, in general, the literature suggests three different kinds of data augmentation, namely easy data augmentation (EDA), back-translation, and mixup. The following subsections discuss the adaptations of the methods appropriate for ABSA and suggest several extensions.

\subsubsection{Easy Data Augmentation for ABSA} \label{sec:eda}
The idea of easy data augmentation (EDA) proposed by \citet{wei2019eda} is to provide an easy and effective way to augment data in NLP. This is done by randomly replacing synonyms, randomly inserting or deleting words, and randomly swapping words. For the case of ABSA, there are some specifics to be considered. First of all, the used model splits the sentence up in left context, target, and right context vectors to determine the polarity, instead of using the sentence as a whole as input. Secondly, the target expression (a word or a group of words) should remain present and unchanged. Removing the target words or splitting the expression up would make a classification of the polarity impossible, as the machine learning algorithm cannot handle this. Keeping these considerations in mind, we propose several adaptations to make EDA compatible with ABSA. Then, we introduce variations and extensions to these methods.

\textbf{Random Insertion}. Random insertion is the process of selecting a random word in the sentence, finding a synonym in WordNet \citep{WordNet} using the natural language toolkit (NLTK) \citep{bird2009natural}, and inserting this synonym at some random place in the sentence. This is possible without many modifications, as this augmentation can be done before splitting the sentence into left, target, and right context. The only thing to keep in mind is that the insertion should not be within the target expression. This is ensured through replacing the target expression with a fixed expression (namely $\$t\$$) before the insertion. Afterward, we replace the fixed expression with the target expression again and append the sentence to the training data. 

\textbf{Random Deletion}. This method selects a random word from a sentence and removes it. For this method, we make sure that the target words cannot be deleted. Therefore, the random deletion procedure is done on the left context vector and the right context vector, not on the target expression. However, we suspect that this process would not work very well in ABSA, as the sentiment of a sentence can easily change when removing a word. If the target, for example, is the food, and the word \textit{delicious} is removed, determining the polarity would be much more difficult for a model.

\textbf{Random Swap}. This method takes two words in the sentence and swaps these words. Again we must make sure that the target words remain unchanged. Replacing the target with a single expression (again $\$t\$$) resolves this issue, as this ensures that the full target expression is swapped instead of only part of the target.  

\textbf{Synonym Replacement}. This method takes a random word in a sentence, looks up a synonym in WordNet, and replaces the word with its synonym. We apply a similar procedure as before and replace the target with a fixed expression ($\$t\$$) to make sure that this expression cannot be selected for the synonym replacement.

These four methods combined create the original EDA for ABSA. All four procedures are similar to a certain extent, so only the synonym replacement is given in Algorithm  \ref{Alg:eda_original}.


\begin{algorithm}[h]
\small
\KwData{$sentences$, sentences of the SemEval dataset; $tar$, target expression in the sentece; $\alpha$, percentage of words replaced}
\KwResult{$augmentedSentences$}
\caption{Synonym replacement\label{Alg:eda_original} in original EDA}
\DontPrintSemicolon
\Begin{
Set $augmentedSentences = \phi$ \;

\ForEach{$originalSentence \in sentences$}{


$sentence \longleftarrow replace(tar,\$t\$,originalSentence)$ \;

$sentence \longleftarrow split(sentence)$\;

$numberAugmentations \longleftarrow \alpha * length(sentence)$ \;
\While{$i < numberAugmentations$}{    
$word \longleftarrow takeRandomWordNotTarget(sentence)$
    
$syn \longleftarrow takeRandomWordNetSynonym(word)$
    
$sentence \longleftarrow $replace $ word $ by $ syn$
    
}
$augmentedSentences \longleftarrow augmentedSentences \cup sentence$\;
}
\Return $augmentedSentences$
}
\end{algorithm}

\subsubsection{Extensions on EDA} \label{sec:eda_adj}
EDA aims to be a quick and straightforward method that can be used on all sorts of NLP tasks. In our case, we look for data augmentation techniques specifically useful for ABSA, and as our training datasets consists of about 1000-2000 sentences, speed is not a big issue for this problem. Therefore, we propose several extensions for EDA.

\textbf{Word Sense Disambiguation}. First of all, synonym replacement and random insertion in the case of EDA pick a random synonym from the WordNet sentiment lexicon. This raises some problems. First of all, when looking up synonyms sometimes words with a different function in a sentence are returned. Take the example in Figure \ref{fig:data_sample}, repeated below in Example 1:

\begin{center}
    \textbf{Example 1.} \textit{Judging from previous posts this used to be a good place, but not any longer.}
\end{center}

When looking up synonyms for the word \textit{judging}, WordNet returns both nouns (like \textit{judgment}) and verbs (like \textit{evaluate}), meaning that the function of a word is context-dependent. Additionally, words with similar functions, or parts-of-speech (POS), can also have different meanings. Take the word \textit{post} in the same sentence. Without context, it can be a military post, mail, or a social media post. This means that many synonyms that are replaced in the original sentences are not actually synonyms, polluting the sentences rather than enhancing them. The process of selecting the correct meaning of the word in a sentence is called word sense disambiguation (WSD) \citep{vasilescu2004evaluating}. 

In order to properly use WSD, we propose a twofold method. First, the POS of every word is determined. This is a complex problem in itself that already got much attention. We use the standard NLTK POS tagger, which is a greedy averaged perceptron tagger \citep{honnibal2013postagger}. 

Afterward, the true sense of the word is determined. An easy algorithm for WSD is proposed by \citet{Lesk1986automatic}. In the Lesk algorithm, all possible definitions of the words in a sentence are looked up and, based on the overlap between two sentences, the proper definition of a word is determined. This method has been refined by the simplified version of Lesk \citep{vasilescu2004evaluating}. Here, a word is looked up in a dictionary, and the definition in the dictionary is used, which gives the highest overlap with the context words. Alternatively, let $w$ be a word for which we want to know its meaning, with its corresponding context words $C = [c_1, c_2, ... , w , ..., c_m]$, which contains $m+1$ words. Let $S = [s_1, s_2, ..., s_n]$ be the possible meanings of the word, with corresponding definitions $[D_1, D_2, ... , D_n]$, where $D_i$ is the set of words in the definitions. We take the meaning with the maximum overlap, where the overlap is calculated as in Equation \ref{eq:or_Lesk}. 

\begin{equation} 
    \label{eq:or_Lesk}
    Overlap(w, s_i) = |D_i \cap C|\\
\end{equation}

The previously discussed method, called simplified Lesk, has been implemented in a WordNet \citep{WordNet} library. WordNet is an extensive lexical database of English words, where nouns, verbs, adjectives, and adverbs are grouped in sets of synonyms called synsets. It attempts to capture all possible meanings of a word with their corresponding definitions, making it a useful tool for WSD. For the simplified Lesk algorithm, the Pywsd library is used \citep{pywsd14}.

Simplified Lesk is used for both synonym replacement and random insertion, adjusting the original implementation of EDA. The pseudocode of synonym replacement in the adjusted manner is given in Algorithm \ref{Alg:eda_adjusted}.

\begin{algorithm}[h]
\small
\KwData{$sentences$, sentences of the SemEval dataset; $tar$, target expression in the sentece; $\alpha$, percentage of words replaced}
\KwResult{$augmentedSentences$}
\caption{Synonym replacement\label{Alg:eda_adjusted} in adjusted EDA}
\DontPrintSemicolon
\Begin{
Set $augmentedSentences = \phi$ \;

\ForEach{$originalSentence \textbf{ in } sentences$}{
$partsOfSpeech \longleftarrow POSTagger(originalSentence)$ \;

$sentence \longleftarrow replace(tar,\$t\$,originalSentence)$    \;

$sentence \longleftarrow split(sentence)$\;

$numberAugmentations \longleftarrow \alpha * length(sentence)$ \;
\While{$i < numberAugmentations$}{
$word \longleftarrow takeRandomWordNotTarget(sentence)$ \;

$maxOverlap \longleftarrow 0$ \;

$bestSense \longleftarrow $most frequent sense of $ word$\;

\ForEach{sense \textbf{in} \textnormal{senses of} word \textnormal{with same part-of-speech in} partsOfSpeech}{
$signature \longleftarrow $set of words in the gloss and examples \;

$overlap \longleftarrow computeWordnetOverlap($

$signature, originalSentence)$

\If{overlap $>$ maxOverlap}{
$maxOverlap \longleftarrow overlap$ \;

$bestSense \longleftarrow sense$

}

}
$sentence \longleftarrow $replace $ word $ with random synonym from the synset corresponding to $ bestSense $ in $ originalSentence$\;

}
$augmentedSentences \longleftarrow augmentedSentences \cup sentence$\;
}
\Return $augmentedSentences$
}
\end{algorithm}

\begin{algorithm}[h]
\small
\KwData{$sentences$, sentences of the SemEval dataset; $asp$, aspect of a given sentence; $categories$, the categories in the SemEval datasets}
\KwResult{$augmentedSentences$}
\caption{Target swap \label{Alg:eda_swap} in adjusted EDA}
\DontPrintSemicolon
\Begin{
Set $augmentedSentences = \phi$ \;
\ForEach{$category \in categories$}{
$sentences \longleftarrow$ sentences in $category$ \;

\For{$sentence_i, sentence_j \in sentences$}{
$sentence_i \longleftarrow replace(asp_i,asp_j, sentence_i)$

$sentence_j \longleftarrow replace(asp_j,asp_i, sentence_j)$

$augmentedSentences \longleftarrow augmentedSentences \cup sentence_i \cup sentence_j$\;
$sentences \longleftarrow sentences - \{sentence_i, sentence_j\}$\;
}
}
\Return $augmentedSentences$
}
\end{algorithm}

\textbf{Target swap across sentences}. In the described tasks, more information is available than only the sentence. In the training data (as can be seen in Figure \ref{fig:data_sample}), a category is given together with its target expression and its position. We can therefore replace random swaps by swapping target words of the same categories. This way, similar targets have multiple contexts in which they can appear. The amount of possible swaps is variable, meaning that it is possible to create many augmented records. We will experiment with one swap per sentence, such that this method can be compared to other data augmentation methods, as this results in the same amount of augmentations. The pseudocode of this method is given in Algorithm \ref{Alg:eda_swap}.

\subsubsection{Backtranslation}
For backtranslation we face a similar problem as with EDA. When translating a sentence to another language and back, the target expression could have changed, resulting in not knowing at which position in the sentence the target is. To tackle this problem, the left context and the right context are independently translated into different languages, while the target remains the same. We pick Dutch and Spanish with Latin alphabets, and we pick Japanese to investigate the effect of translation to a non-Latin alphabet. In order to translate the data and back the Google Translate API is used\footnote{https://cloud.google.com/translate/docs/basic/translating-text}. Algorithm \ref{Alg:backtranslation} describes the way this backtranslation is done.

\begin{algorithm}[h]
\small
\KwData{$sentences$, sentences of the SemEval dataset; $tar$, target expression in a sentence; $lang$, target language to translate to}
\KwResult{$augmentedSentences$}
\caption{Backtranslation \label{Alg:backtranslation}}
\DontPrintSemicolon
\Begin{
Set $augmentedSentences = \phi$ \;

\ForEach{$originalSentence \in sentences$}{

$left, target, right \longleftarrow split(originalSentence, tar)$

$translatedLeft \longleftarrow translate(left, lang)$\;

$translatedRight \longleftarrow translate(right, lang)$\;

$backtranslatedLeft \longleftarrow translate(translatedLeft, Eglish)$ \;

$backtranslatedRight \longleftarrow translate(translatedRight, English)$\;

$backtranslated \longleftarrow backtranslatedLeft + target + backtranslatedRight$\;

$augmentedSentences \longleftarrow augmentedSentences \cup backtranslated$\;
}
\Return $augmentedSentences$
}
\end{algorithm}

\subsubsection{Mixup}
Mixup is a linear interpolation between feature vectors, identical to the interpolation of the associated class vectors \citep{zhang2017mixup}. The technique of mixup is used initially for image classification tasks, but recently had been applied for NLP. For this paper, we introduce a variant of mixup appropriate for the considered two-step hybrid model for ABSA.

In order to properly apply mixup at a sentence level, \citet{ghou2019mixup} introduced word-mixup. This way the input sentences are zero-padded on the right in order to make them of the same length, after which the linear interpolation is done. As our model makes use of a left context, target, and right context, we propose the following method. We define $S_i = [w_1, w_2, ..., w_{N_i}]$ as a sentence $i$ which contains $N_i$ words, and $S_j = [w_1, w_2, ... , w_{N_j}]$ as sentence $j$ with $N_j$ words. We first split the sentence up in $S_{i,l} = [l_1, l_2, ..., l_{L_i}]$, $ S_{i,c} = [c_1, c_2, ..., c_{C_i}]$, and $S_{i,r} = [r_1, r_2, ..., s_{R_i}]$, which are the left context, target, and right context, respectively ($L_i + C_i + R_i = N_i$). The polarity vector is a one hot encoded vector of dimension three, meaning that $y_i \in \mathcal{R}^3$. We perform similar operations on all three vectors, so we take the left vector as an example. The first step is to apply the right zero padding for the left context. This creates $[l_1, l_2, ..., l_{L_i}, o_1, o_2, ..., o_{Q_l-L_i}]$ where $Q_l$ is the maximum length of the left context vector and $o_i \in \mathcal{R}^d$ is an embedding with zeros ($d$ is the embedding dimension). We obtain $S_{i,l} \in \mathcal{R}^{d \times Q_l}$. Afterwards, the following linear interpolation is done:

\begin{align}
    \label{eq:mixup}
     \tilde{S}_{ijl} & = \lambda S_{i,l} + (1-\lambda) S_{j,l} \\
    \tilde{y}_{ij} & = \lambda y_i + (1-\lambda) y_j
\end{align}

\noindent where $\tilde{S}_{ijl}$ is the left context vector of the augmented record, $ \tilde{y}_{ij}$ is the polarity of the augmented record, and $\lambda$ is a random value drawn from the distribution $Beta(\alpha, \alpha)$. Mixup regularizes the NN such that the linear behavior is preferred to more complex functions. This reduces overfitting when training, and makes the training more robust when having corrupt labels. \citet{zhang2017mixup} found that an $\alpha \in [0.1, 0.4]$ worked well, as values higher than this resulted in underfitting. This is because linearly interpolating with a low $\alpha$ results in a high probability that $\lambda$ takes a value in the tails (so either close to zero or close to one). This way, the augmented data is slightly interpolated, which should prevent overfitting, but not so much that it underfits. In this paper we will therefore experiment with $\alpha \in [0.1, 0.4]$.

\section{Evaluation}\label{sec:evaluation}
This section presents the best results of the performed experiments. Section \ref{sec:baseline} presents the baseline method. Section \ref{sec:eda_res} gives the results of the augmentation techniques. Last, Section \ref{sec:combinations} discusses some insights in the obtained results.


\subsection{Baseline and Comparison} \label{sec:baseline}
While reproducing the results presented in \citet{HAABSA}, we found that the ontology reasoner has an accuracy of $0.87$ for the data of 2016, similar to the reported accuracy, and $0.83$ for the dataset of 2015, which is slightly higher than the reported accuracy. However, as our paper aims to investigate the effect of data augmentation, this is not troublesome. The reproduced model is used as a baseline, and the accuracy of the predictions on the test set of SemEval 2015 and SemEval 2016 is the employed performance measure. 


\subsection{Augmentation techniques} \label{sec:eda_res}
In Table \ref{tab:EDA_total}, the baseline is compared to the individual results of the augmentation techniques. While the EDA stands for the combined solution with the four methods presented in Section \ref{sec:eda} (random insertion, random deletion, random swap and synonym replacement), adjusted EDA represents the techniques introduced in Section \ref{sec:eda_adj} (random insertion and synonym replacement with word sense disambiguation, and target swap). In terms of both SemEval datasets, it can be seen that the adjusted EDA outperforms the original EDA. Both EDA approaches lead to a performance boost with respect to the baseline HAABSA between 0.3-1 and 0.2-0.5 (EDA-adjusted EDA) percentage points of accuracy for SemEval 2015 and SemEval 2016 test datasets, respectively. 

Furthermore, the backtranslation technique is assessed using Dutch, Spanish, and Japanese. This method has varying results, sometimes increasing or decreasing accuracy for different languages. We suspect that backtranslation generates varying results because the target expression had to remain fixed. This way, the augmentation techniques were performed on a part of a sentence instead of on a sentence as a whole. For backtranslation, we translate the left and the right contexts to the target language and back, keeping the target fixed. This way, the translation engine has to translate different parts of a sentence. While the backtranslation from Dutch slightly outperforms the other languages and the baseline with $0.1$ percentage points for the case of SemEval 2015 test dataset, the backtranslation from Japanese improve the accuracy of SemEval 2016 test dataset till the level of adjusted EDA. As a result, we consider that the Japanese language yields the best results for the backtranslation. 

The last data augmentation technique we evaluate is the mixup with $\alpha \in [0.1, 0.4]$. We notice that the setting of the $\alpha$ parameter to $0.2$ produces the best accuracy, improving the accuracy of HAABSA applied on the SemEval 2016 test data with $0.4$ percentage points. On the contrary, the accuracy on the SemEval 2015 test data stays constant or slightly decreases ($\alpha =0.1$). For mixup, we had to linearly interpolate between two different left contexts, two different targets, and two different right contexts. As sentences can have very different structures, linear interpolation between parts of sentences appears to have a small effect on the accuracy. 

We empirically found that the 1:1 ratio of augmentations and original records yield the highest results, after we compared a 3:1 ratio and a 1:3 ratio. Overall, we can conclude that the best technique is the adjusted EDA. The largest boost in performance is given by the target swap, followed by adjusted synonym replacement, and then adjusted random insertion. However, given that the accuracy of HAABSA also takes into account the ontology, the effect of the adjusted EDA technique, as it is shown in Table \ref{tab:EDA_total}, is diminished. Knowing that the accuracy of the baseline multi-hop LCR-Rot is $72.9\%$ and $78.9\%$ on the SemEval 2015 and SemEval 2016 test data, respectively, the adjusted EDA leads to an increase of $1.3$ percentage points on the SemEval 2016 dataset, and an increase of $2$ percentage points on the SemEval 2015 dataset. 


\begin{table*}[hbt!] 
\caption{Results for different data augmentation methods. Best results per technique are given in bold.}
\label{tab:EDA_total}
\centering
\small
\begin{tabular}{l|lll|lll} 

\multirow{2}{*}{{Model}} & \multicolumn{3}{c}{2015} & \multicolumn{3}{c}{2016}  \\
            & training acc. & testing acc. & \#aug. & training acc. & testing acc.& \#aug.\\ 
\hline
HAABSA        & 90.9 \% & 77.9\% & 0 & 85.7\% & 83.9\% & 0       \\
\hline
HAABSA + EDA & 85.8\% & 78.2\% & 5116 & 96.2\% &84.1\% & 7520       \\
HAABSA + EDA adj. & \textbf{95.2\%} &\textbf{78.9\%} & 3837 & \textbf{96.5\%} &\textbf{84.4}\% & 5640      \\
\hline
HAABSA + BT NL & 89.9\% & 78.0\% & 1278 & 78.9\% & 83.8\% & 1880 \\
HAABSA + BT ES & 85.1\% & 77.6\% & 1278 & 82.0\% & 84.3\% & 1880 \\
HAABSA + BT JA  & 86.9\% & 77.9\% & 1278 & 81.3\%  & \textbf{84.4\%} & 1880 \\
\hline
HAABSA mixup $\alpha =0.1$ & 88.0\% & 77.6\% & 1278 & 80.8\% & 83.8\% & 1880       \\
HAABSA mixup $\alpha =0.2$ & 85.0 \% & 77.9\% & 1278 & 80.4\% & 84.3\% & 1880       \\
HAABSA mixup $\alpha =0.3$ & 89.2\% & 77.9\% & 1278 & 83.8\% & 84.1\% & 1880       \\
HAABSA mixup $\alpha =0.4$ & 82.8\% & 77.9\% & 1278 & 79.6\% & 84.1\% & 1880       \\

\end{tabular}
\end{table*}

\begin{table*}[hbt!] 
\caption{Results of EDA adjusted ratio variations Best results per technique are given in bold.}
\label{tab:EDA_ratio}
\centering
\small
\begin{tabular}{l|lll|lll} 

\multirow{2}{*}{{Model}} & \multicolumn{3}{c}{2015} & \multicolumn{3}{c}{2016}  \\
& training acc. & testing acc. & \#aug. & training acc. & testing acc.& \#aug.\\
\hline

HAABSA + EDA adj. 1:1 & \textbf{95.2\%} &\textbf{78.9\%} & 3837 & \textbf{96.5\%} &\textbf{84.4}\% & 5640 \\
HAABSA + EDA adj. 1:3 & 86.1\% & 78.1\% & 3834 & 80.7\% & 84.1\% & 5640 \\
HAABSA + EDA adj. 3:1 & 84.4\% & 77.9\% & 3834 & 91.1\% & 83.5\% & 5640 \\

\end{tabular}
\end{table*}

\subsection{Insights} \label{sec:combinations}
To understand why some of data augmentations techniques are more effective than others, we need to assess their capacity to generate meaningful sentences. In doing this, we randomly select two sentences from the employed datasets. Random deletion and random insertion are left out, as these follow logically from the given examples, similarly to the backtranslation of other sentences. Additionally, the categories for both sentences are the same, namely $SERVICE\#GENERAL$, such that the target swap can be performed. As can be seen from Table \ref{tab:examples}, the adjusted EDA methods (synonym replacement and target swap) mainly have the desired effect of creating new sentences with extra information. This inference straightens the idea already suggested in Table \ref{tab:EDA_total}, according to which adjusted EDA is the most effective data augmentation technique. On the other hand, the simple EDA methods usually seem to pollute the data in a way that does not add extra information or adds wrong information. For example, when the algorithm replaces the word ``us" with the atomic number ``92". This is because ``us" can be interpreted as the plural of ``u", where ``u" is the atomic symbol for Uranium. 


\begin{table}[]
    \caption{Individual examples of augmentations, aspects are in bold}
    \label{tab:examples}
    \small
    \begin{tabularx}{\linewidth}{
        >{\hsize=.22\hsize}X
        |>{\hsize=0.4\hsize}X
        >{\hsize=0.4\hsize}X
      } 
        \textbf{Type augmentation} & \textbf{Sentence 1} & \textbf{Sentence 2} \tabularnewline
       \hline
        Original & the \textbf{hostess} is rude to the point of being offensive & The \textbf{waitress} was very patient with us and the food is phenomenal! \tabularnewline
        \hline
        EDA original random swap & \textit{being} \textbf{hostess} is rude to  point of \textit{the} offensive & The \textbf{waitress} \textit{food} very with patient us and the \textit{was} is phenomenal! \tabularnewline
        \hline
        EDA original synonym replacement & the \textbf{hostess} is rude to the \textit{breaker point} of being nauseous & The \textbf{waitress} was very patient with \textit{atomic number 92} and the food is phenomenal! \tabularnewline
        \hline
        Adjusted EDA synonym replacement & the \textbf{hostess} is \textit{uncivil} to the point of being \textit{unsavory} & The \textbf{waitress} was very patient with us and the \textit{nourishment} is phenomenal!  \tabularnewline
        \hline
        Adjusted EDA Target Swap & the \textbf{waitress} is rude to the point of being offensive & The \textbf{hostess} was very patient with us and the food is phenomenal! \tabularnewline
        \hline
        Backtranslation Japanese & of \textbf{hostess} it’s rude enough to cause discomfort.   & of \textbf{waitress} very patient and the food is amazing! \tabularnewline
      \end{tabularx}
\end{table}

Another interesting observation is the way backtranslation works. In Table \ref{tab:examples}, the word ``the", when translated to Japanese and back, becomes ``of". Additionally, when translating a part of a sentence to Japanese and back, the translation engine attempts to create sentences from only part of a sentence. Therefore, the context on the right of both sentences can be read as a standalone sentence. Even if the meaning of the initial sentence is usually kept, sometimes, the backtranslation process might negatively influence the overall accuracy. 

Lastly, the ratio of original sentences and augmented sentences has been modified to 3:1, 1:1 and 1:3. This means that the original data is used thrice for the 3:1, and that the augmentations are done thrice in the 1:3. However, this experiment showed that a 1:1 ratio yielded the best results, as it is shown in Table \ref{tab:EDA_ratio}.

\section{Conclusion} \label{sec:concludingremarks}

This paper focuses on data augmentation methods in the field of ABSA on a sentence level by extending the state-of-the-art model proposed by \citet{HAABSA}. The first proposed approach is called EDA, and it is inspired by the work of \citet{wei2019eda}, who created a general set of four methods applicable for all NLP classification tasks. 
As this approach proves to be too naive to properly address the ABSA task, some adjustments are introduced based on word sense disambiguation and swaps between targets that share the same category. Besides the EDA methods and their variations, we also investigated the effect of the backtranslation and the mixup methods as a linear interpolation between feature vectors. Both methods only improve the HAABSA applied on SemEval 2016 test data, while generating similar results with the baseline on the SemEval 2015 data. Overall, the adjusted EDA is the most effective data augmentation method, increasing the accuracy of the SemEval 2015 and SemEval 2016 test datasets with $1.0$ and $0.5$ percentage points, respectively.

Furthermore, in this paper, the data augmentation techniques were used on all sentences. It would be interesting to analyze which kind of sentences give the highest yield when being augmented and which sentences harm the accuracy when being augmented. One direction to consider is the effect of the length of a sentence on the data augmentation. This analysis can also be extended for mixup, in the sense that instead of selecting random sentences used for mixup, it might be useful to select only the sentences with similar lengths for their contexts. This implies that the augmentation techniques do not necessarily have to be used upon all sentences, but can be used only on the ones with similar characteristics.

\bibliographystyle{ACM-Reference-Format}
\balance
\bibliography{bibliography}

\end{document}